\documentclass[11pt,a4paper]{article}
\usepackage[hyperref]{emnlp2020}
\usepackage{times}
\usepackage{latexsym}

\usepackage{microtype}

\aclfinalcopy 

\usepackage{graphicx}
\usepackage{subfig}

\usepackage{multirow}
\usepackage{amssymb}
\usepackage{pifont}
\newcommand{\xmark}{\ding{55}}%
\usepackage{amsmath}
\allowdisplaybreaks
\usepackage{array}

\DeclareMathAlphabet{\mathcal}{OMS}{cmsy}{m}{n}
\definecolor{mypink}{rgb}{1., 0, 0}
\usepackage{tabularx}

\title{GRACE: Gradient Harmonized and Cascaded Labeling for Aspect-based Sentiment Analysis}

\author{Huaishao Luo$^1$\thanks{~~Work is done during an internship at MSR Asia.}, ~Lei Ji$^{2,4,5}$, Tianrui Li$^1$\thanks{~~Correspongding author.}, ~Nan Duan$^{2}$, Daxin Jiang$^{3}$\\
    $^1$School of Information Science and Technology, Southwest Jiaotong University, China\\
    {\tt huaishaoluo@gmail.com, trli@swjtu.edu.cn}\\
	$^2$Microsoft Research Asia, Beijing, China\\
	$^3$STCA NLP Group, Microsoft, Beijing, China \\
	$^4$Institute of Computing Technology, Chinese Academy of Science, Beijing, China\\
	$^5$University of Chinese Academy of Sciences, Beijing, China \\
	{\tt \{leiji,nanduan,djiang\}@microsoft.com}
}

\date{}
\begin{document}
\maketitle
\begin{abstract}
In this paper, we focus on the imbalance issue, which is rarely studied in aspect term extraction and aspect sentiment classification when regarding them as sequence labeling tasks. Besides, previous works usually ignore the interaction between aspect terms when labeling polarities. We propose a \textbf{GR}adient h\textbf{A}rmonized and \textbf{C}ascad\textbf{E}d labeling model (GRACE) to solve these problems. Specifically, a cascaded labeling module is developed to enhance the interchange between aspect terms and improve the attention of sentiment tokens when labeling sentiment polarities. The polarities sequence is designed to depend on the generated aspect terms labels. To alleviate the imbalance issue, we extend the gradient harmonized mechanism used in object detection to the aspect-based sentiment analysis by adjusting the weight of each label dynamically. The proposed GRACE adopts a post-pretraining BERT as its backbone. Experimental results demonstrate that the proposed model achieves consistency improvement on multiple benchmark datasets and generates state-of-the-art results.
\end{abstract}

\section{Introduction}
\label{sec_introduction}
\begin{figure}[htb]
	\centering
	\includegraphics[width=0.40\textwidth, keepaspectratio]{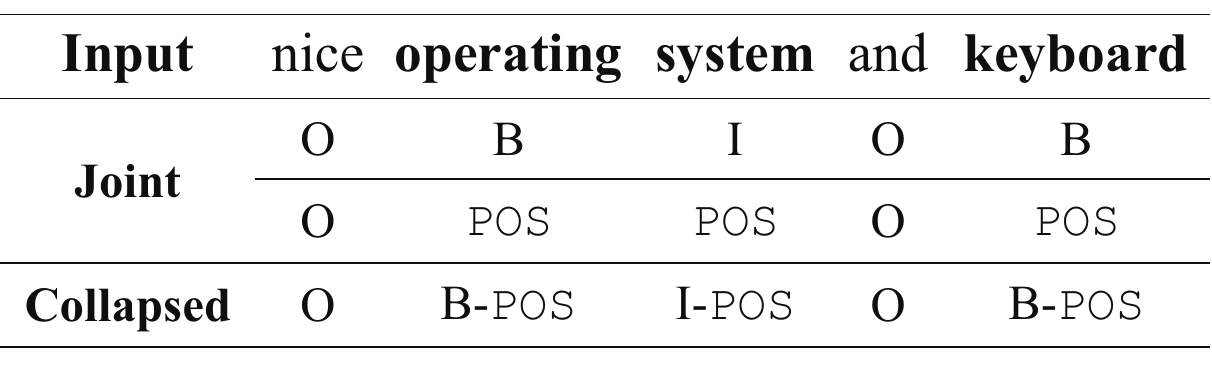}
    \caption{Joint and Collapsed labeling approaches on aspect terms and their polarities. \texttt{POS} means positive.}
    \label{table_labeling_examples}
\end{figure}

Aspect terms extraction (ATE) and aspect sentiment classification (ASC) are two fundamental, fine-grained subtasks in aspect-based sentiment analysis (ABSA). ATE is the task of extracting the aspect terms (or attributes) of an entity upon which opinions have been expressed, and ASC is the task of identifying the polarities expressed on these extracted terms in the opinion text \cite{Hu2004}. Consider the example in Figure \ref{table_labeling_examples}, which contains comments that people expressed about the aspect terms ``operating system" and ``keyboard", and their polarities are all positive.
\begin{figure}[htb]
    \centering
    \subfloat[Label statistics] {
        \centering
        \includegraphics[width=0.81in]{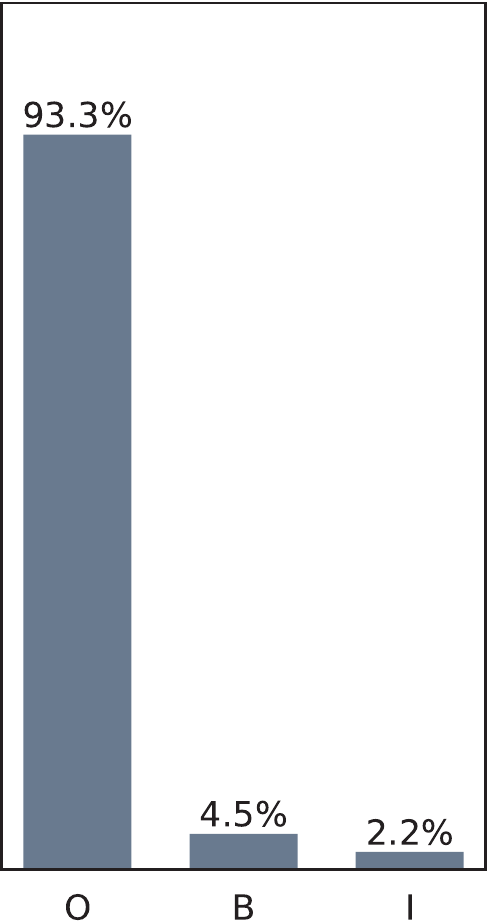}
        \label{fig_imbalanceStatistic}
    }
    \subfloat[Gradient statistics] {
        \centering
        \includegraphics[width=1.532in]{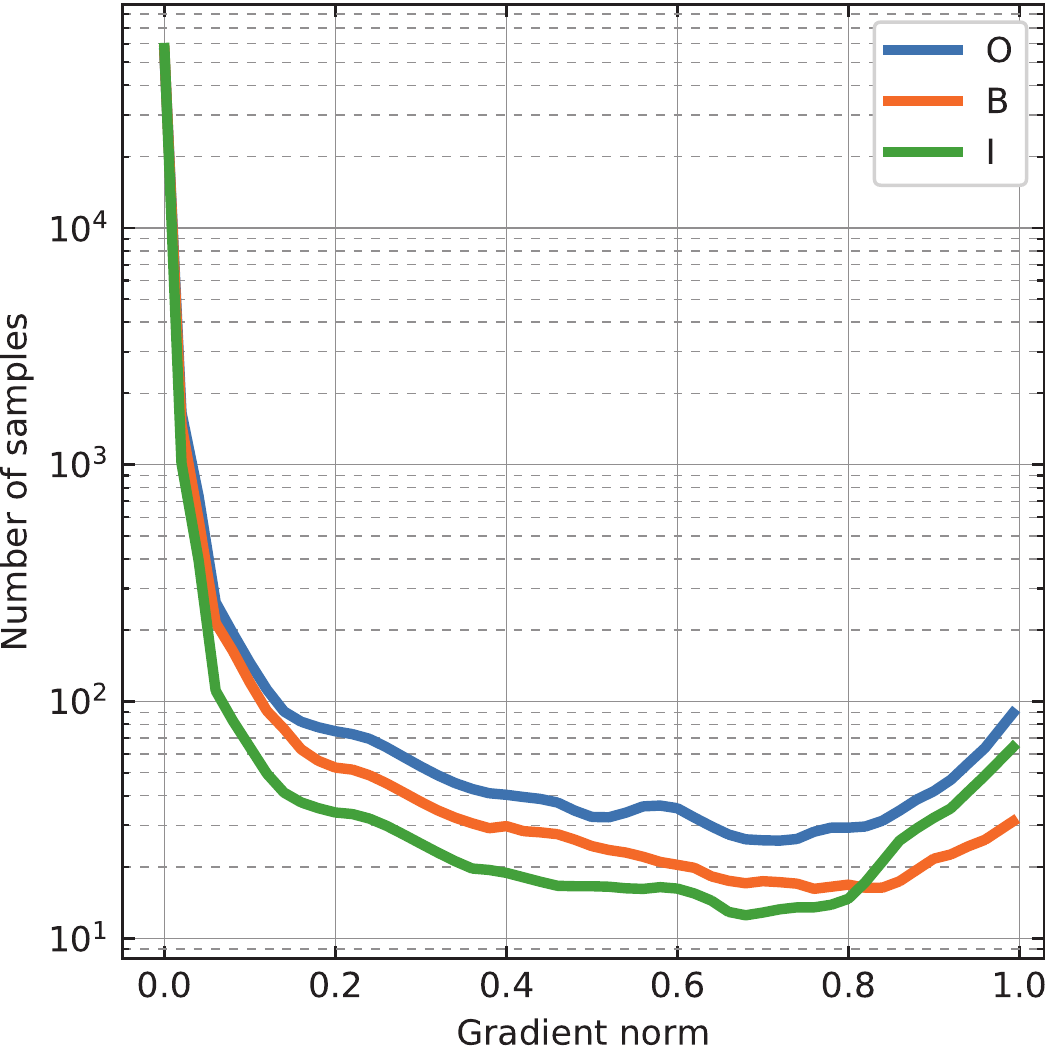} 
        \label{fig_gradientHarmonized}
    }
    \caption{Label statistics and gradient distribution on the laptop dataset of SemEval-14. The y-axis in (b) uses a log scale.}  
    \label{fig_statistic}
\end{figure}

To better satisfy the practical applications, the aspect term-polarity co-extraction, which solves ATE and ASC simultaneously, receives much attention in recent years \cite{Li2019a,Luo2019,Hu2019,Wan2020}. A big challenge of the aspect term-polarity co-extraction in a unified model is that ATE and ASC belong to different tasks: ATE is usually a sequence labeling task, and ASC is usually a classification task. Previous works usually transform the ASC task into sequence labeling. Thus the ATE and ASC have the same formulation.

There are two approaches of sequence labeling on the aspect term-polarity co-extraction. As shown in Figure \ref{table_labeling_examples}, one is the joint approach, and the other is the collapsed approach. The preceding one jointly labels each sentence with two different tag sets: aspect term tags and polarity tags. The subsequent one uses collapsed labels as the tags set, e.g., ``B-POS'' and ``I-POS'', in which each tag indicates the aspect term boundary and its polarity. Except for the joint and collapsed approaches, a pipelined approach first labels the given sentence using aspect term tags, e.g., ``B'' and ``I'' (the beginning and inside of an aspect term), and then feeds the aspect terms into a classifier to obtain their corresponding polarities.

Several related works have been published in these approaches. \citet{Mitchell2013} and \citet{Zhang2015} found that the joint and collapsed approaches are superior to the pipelined approach on named entities and their sentiments co-extraction. \citet{Li2019a} proposed a unified model with the collapsed approach to do aspect term-polarity co-extraction. \citet{Hu2019} solved this task with a pipelined approach. \citet{Luo2019} adopted the joint approach to do such a co-extraction. We follow the joint approach in this paper, and believe that it has a more apparent of responsibilities than the collapsed approach through learning parallel sequence labels.

However, previous works on the joint approach usually ignore the interaction between aspect terms when labeling polarities. Such an interaction is useful in identifying the polarity. As an instance, in Figure \ref{table_labeling_examples}, if ``operating system'' is positive, ``keyboard'' should be positive due to these two aspect terms are connected by coordinating conjunction ``and''. Besides, almost all of previous works do not concern the imbalance of labels in such sequence labeling tasks. As shown in \ref{fig_imbalanceStatistic}, the number of `O' labels is much larger than that of `B' and `I', which tends to dominant the training loss. Moreover, we find the same gradient phenomenon as \citet{Li2019} in the sequence labeling task. As shown in Figure \ref{fig_gradientHarmonized}, most of the labels own low gradients, which have a significant impact on the global gradient due to their large number.

\begin{figure*}[tp] 
    \centering
    \includegraphics[width=0.93\textwidth]{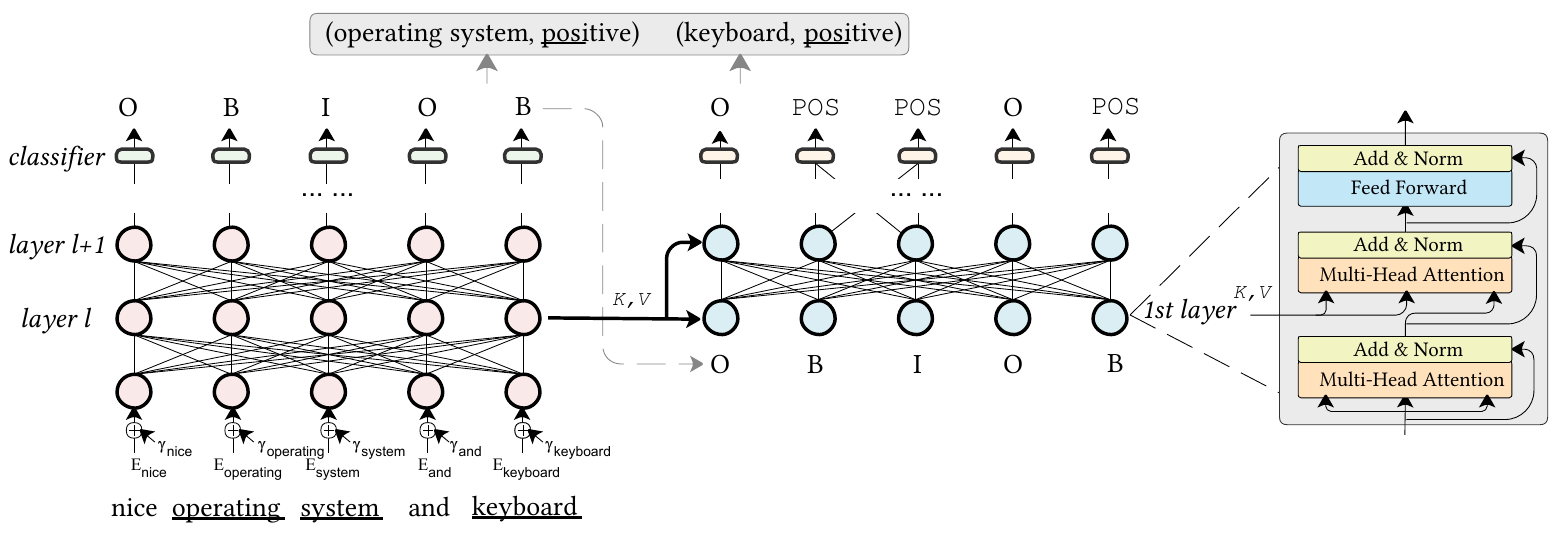} 
    \caption{The main structure of our \textbf{GRACE}. It is a cascaded labeling architecture, which means that the generated aspect term labels [O,B,I,O,B] are fed to the right part as key $K$ and value $V$ to generate sentiment labels [O,\texttt{POS},\texttt{POS},O,\texttt{POS}]. The perturbed embeddings $r_\cdot$ is added to the Token embeddings $\mathit{E}_\cdot$.}
    \label{fig_main_structure}
\end{figure*}

Considering the above issues, we propose a \textbf{GR}adient h\textbf{A}rmonized and \textbf{C}ascad\textbf{E}d labeling model (\textbf{GRACE}) in this paper. The proposed GRACE is shown in Figure \ref{fig_main_structure}. Unlike previous works, GRACE is a cascaded labeling model, which uses the generated aspect term labels to enhance the polarity labeling in a unified framework. Specifically, we use two encoder modules shared with lower layers to extract representation. One encoder module is for ATE, and the other is for ASC after giving the aspect term labels generated by the preceding encoder. Thus, the GRACE could consider the interaction between aspect terms in the ASC module through a stacked Multi-Head Attention \cite{Vaswani2017}. Besides, we extend a gradient harmonized loss to address the imbalance labels in the model training phase.

Our contributions are summarized as follows:
\begin{itemize}
    \item A novel framework GRACE is proposed to address the aspect term-polarity co-extraction problem in an end-to-end fashion. It utilizes a cascaded labeling approach to consider the interaction between aspect terms when labeling their sentiment tags.
    \item The imbalance issue of labels is considered, and a gradient harmonized strategy is extended to alleviate it. We also use virtual adversarial training and post-training on domain datasets to improve co-extraction performance.
\end{itemize}
In the following, we describe the proposed framework GRACE in Section \ref{sec_model}. The experiments are conducted in Section \ref{sec_experiments}, followed by the related work in Section \ref{sec_related_work}. Finally, we conclude the paper in Section \ref{sec_conclusion}.

\section{Model}
\label{sec_model}
An overview of GRACE is given in Figure \ref{fig_main_structure}. It is comprised of two modules with the shared shallow layers: one is for ATE, and the other is for ASC. We will first formulate the co-extraction problem and then describe the framework in detail in this section.

\subsection{Problem Statement}
This paper deals with aspect term-polarity co-extraction, in which the aspect terms are explicitly mentioned in the text. We solve it as two sequence labeling tasks. Formally, given a review sentence $S$ with $n$ words from a particular domain, denoted by $S=\{w_i | i = 1, \dots, n\}$. For each word $w_i$, the objective of our task is to assign a tag $t^e_i \in T^e$, and a tag $t^c_i \in T^c$ to it, where $T^e=\{$B, I, O$\}$ and $T^c=\{$\texttt{POS}, \texttt{NEU}, \texttt{NEG}, \texttt{CON}, O$\}$. The tags `B', `I' and `O' in $T^e$ stand for the beginning, the inside of an aspect term, and other words, respectively. The tags \texttt{POS}, \texttt{NEU}, \texttt{NEG}, and \texttt{CON} indicate polarity categories: positive, neutral, negative, and conflict, respectively~\footnote{We regard \textit{neutral} as a polarity as many prior works.}. The tag `O' in $T^c$ means other words like that in $T^e$. Figure \ref{table_labeling_examples} shows a labeling example of the joint and collapsed approaches.

\subsection{GRACE: Gradient Harmonized and Cascaded Model}
\label{sec_grace_mdoel}
We focus on the joint labeling approach in the paper. As shown in Figure \ref{fig_main_structure}, the proposed GRACE contains two branches with the shared shallow layers. In order to benefit from the pretrained model, we use the BERT-Base as our backbone. Then the representation $\mathbf{H}_{e}$ of ATE can be generated on the pretrained BERT:
\begin{align}
    &\mathbf{H}^{[1:L]} = \text{BERT}(S), \\
    &\mathbf{H}_{e} = \mathbf{H}^{L}, \label{eq_bert_ate}
\end{align}
where $\mathbf{H}^{[1:L]}$ denotes the representation of each layer of BERT. It varies from the 1st layer to the $L$-th layer. $L$ is the max layer of BERT, e.g., 12 in BERT-Base.  $\mathbf{H}_{e} \in \mathbb{R}^{(\hat{n}+2) \times h}$ is the representation $\mathbf{H}^{L}$ belonging to the last layer, in which two extra embeddings belong to special tokens [CLS] and [SEP], and the labels of them are set to `O' in the experiments. $h$ is the hidden size, $\hat{n}$ is the length of $S$ after tokenizing by the wordpiece vocabulary.

Different layers of BERT capture different levels of information, e.g., phrase-level information in the lower layers and linguistic information in intermediate layers \cite{Jawahar2019}. The higher layers are usually task-related. Thus, a shared BERT between ATE and ASC tasks is the right choice. We extract the representation $\mathbf{H}_{c}$ for ASC task from the $l$-th layer of BERT:
\begin{align}
	&\mathbf{H}_{c} = \mathbf{H}^{l}. \label{eq_bert_asc}
\end{align}
Thus, $\mathbf{H}^{[l+1:L]}$ is task-specific for ATE. An extreme state is $l=L$, where all layers are shared across both tasks. We omit an exhaustive description of BERT and refer readers to \citet{Devlin2019} for more details.

\noindent
\textbf{Cascaded Labeling}\quad
We can do sequence labeling on the $\mathbf{H}_{e}$ and $\mathbf{H}_{c}$ directly. However, it is not a customized feature for ASC. Conversely, ASC may decline the ATE performance. One reason is the difference between ATE and ASC. The polarity of an aspect term usually does not come from the term itself. For example, the polarity of aspect term ``operating system'' in Figure \ref{table_labeling_examples} comes from the adjective ``nice''. When labeling the ``operating system'', the model needs to point to the ``nice''. The other reason is the interaction between aspect terms is ignored when labeling their sentiment labels. For example, the ``operating system'' and ``keyboard'' are connected by coordinating conjunction ``and''. If ``operating system'' is positive, ``keyboard'' should be positive, too.

Thus, we propose the cascaded labeling approach, which uses the generated aspect terms sequence as the input to generate the sentiment sequence. As shown in Figure \ref{fig_main_structure}, the $\mathbf{H}_{c}$ is fed to a new Transformer-Decoder \cite{Vaswani2017} as key $K$ and value $V$ to generate a new aspect sentiment representation $\mathbf{G}_{c}$:
\begin{align}
	\mathbf{G}_{c} = \text{Transformer-Decoder}(\mathbf{H}_{c}, Q), \label{eq_grace_asc}
\end{align} 
where $Q$ represents the aspect term labels generated by the ATE module (ground-truth labels in the training phase). The vocab size is $|T^e|$ in the word embedding of the Transformer-Decoder. 

Note that the Transformer-Decoder here is not the same as the original transformer decoder. The difference is that we use Multi-Head Attention instead of Masked Multi-Head Attention as the first sub-layer because the ASC is not an autoregressive task and does not need to predict the output sequence one element at a time.

\noindent
\textbf{Gradient Harmonized Loss}\quad
The cross entropy is used to train the model:
\begin{align}
	&p^\tau= \textrm{softmax}(\mathbf{M}_{\tau}\mathbf{w}_\tau), \label{eq_softmax_pro} \\
	&\mathcal{L}_\tau = -\frac{1}{n}\sum_{i=1}^n \left( \mathbb{I} \left(t_i^\tau \right) \left( \text{log} \left(p^\tau_i \right) \right)^\top \right), \label{eq_each_loss}
\end{align}
where $\tau \in \{e,c\}$, $\mathbf{M}=\mathbf{H}$ if $\tau$ is $e$, and $\mathbf{M}=\mathbf{G}$ if $\tau$ is $c$, $\mathbb{I}(t_i^\tau)$ means the one-hot vector of $t_i^\tau \in T^\tau$, $\mathbf{w}_\tau$ is a trainable weight matrix. 

Then, the losses from both tasks are constructed as the joint loss of the entire model:
\begin{align}
    \mathcal{J}(\Theta) = \mathcal{L}_e + \mathcal{L}_c,
\end{align}
where $\mathcal{L}_e$ and $\mathcal{L}_c$ denote the loss for aspect term and polarity, respectively. $\Theta$ represents the model parameters containing all trainable weight matrices and bias vectors. 

However, there are two well-known disharmonies to affect the performance through the optimization of the above losses. The first one is the imbalance between positive and negative examples, and the other one is the imbalance between easy and hard examples \cite{Li2019}. Specifically, there exists the imbalance between each label in our labeling task. As shown in Figure \ref{fig_imbalanceStatistic}, the label `O' occupies a tremendous rate than other labels. According to the work from \citet{Li2019}, the easy and hard attributes of labels can be represented by the norm of gradient $g$:
\begin{align}
    g= \left| \frac{\partial \mathcal{L}}{\partial z} \right| = \vert p - \hat{t} \vert, \label{eq_cal_gradient}
\end{align}
where $\hat{t}$ is the ground-truth with value 0 or 1, $p$ is the score calculated by a softmax operation, $z$ is the logit output of a model, $\mathcal{L}$ is the cross entropy. E.g., $z=\mathbf{M}_{\tau}\mathbf{w}_\tau$ and $p$ in Eq. (\ref{eq_softmax_pro}), and $\mathcal{L}$ in Eq. (\ref{eq_each_loss}).

Figure \ref{fig_gradientHarmonized} shows the statistic of labels w.r.t gradient norm $g$. Most of the labels own low gradients, which have a significant impact on the global gradient due to their large number. A strategy is to decrease the weight of loss from these labels.

We rewrite the Eq. (\ref{eq_each_loss}) following GHM-C, which used in object detection \cite{Li2019}, as follows:
\begin{align}
    &\mathcal{L}_\tau = -\frac{1}{n}\sum_{i=1}^n \left( \beta_{t_i^\tau} \big( \mathbb{I} \left(t_i^\tau \right) \left( \text{log} \left(p^\tau_i \right) \right)^\top \big) \right), \label{eq_each_weighted_loss} \\
    &\beta_{t_i^\tau} = \frac{N^\tau}{\rho\left(g_{t_i^\tau}\right)},
\end{align}
where $g_{t_i^\tau}$ is the gradient norm of $t_i^\tau$ calculated by Eq. (\ref{eq_cal_gradient}), $N^\tau$ is the total number of labels, $\rho(g)$ is gradient density:
\begin{align}
    &\rho(g)=\frac{1}{l_{\epsilon}(g)} \sum_{k=1}^{N^\tau} \delta_{\epsilon}\left(g_{k}, g\right), \label{eq_gradient_density}
\end{align}
where $\delta_{\epsilon}(x, y)$ is 1 if $y-\frac{\epsilon}{2}\leq x<y+\frac{\epsilon}{2}$ otherwise 0. The $\rho(g)$ denotes the number of labels lying in the region centered at $g$ with a length of $\epsilon$ and normalized by the valid length $l_{\epsilon}(g)=\min \left(g+\frac{\epsilon}{2}, 1\right)-\max \left(g-\frac{\epsilon}{2}, 0\right)$ of the region.

The calculation of $\beta_{t_i^\tau}$ will use the unit region to reduce complexity. Specifically, the gradient norm $g$ will put into $m={1}/{\epsilon}$ unit regions. For the $j$-th unit region $u_j=\big[j\epsilon-\epsilon, j\epsilon\big)$, the gradient density can be approximated as:
\begin{align}
    \hat{\rho}(g)=U_{ind(g)}/{\epsilon}=mU_{ind(g)}, 
\end{align}
where $U_j$ denotes the number of labels lying in $u_j$, $ind(g)=\kappa$ s.t. $(\kappa-1)\epsilon \leq g<\kappa\epsilon$ is the index of the unit region in which $g$ lies.

The calculation of $\hat{\rho}(g)$ assumpts that the examples lying in the same unit region share the same gradient density. So it
can be calculated by the algorithm of histogram statistics.

A further reasonable manner is to statistics $U_j^{(t)}$ in the $t$-th iteration to reduce the complexity of $U_j$'s statistic cross the dataset, and uses $A_j^{(t)}$ to approximate the real $U_j$ as follows:
\begin{align}
    A_j^{(t)}=\alpha A_j^{(t-1)} + (1-\alpha) U_j^{(t)}, \label{eq_momentum}
\end{align}
where $\alpha$ is a momentum parameter. Thus, the $\hat{\rho}(g)$ is updated by:
\begin{align}
    \hat{\rho}(g)={A_{ind(g)}} / {\epsilon}=mA_{ind(g)}, 
\end{align}

\noindent
\textbf{Virtual Adversarial Training}\quad 
To make the model more robust to adversarial noise, we utilize the virtual adversarial training (VAT) used in \cite{Miyato2016} to make small perturbations $r$ to the input Token Embedding $E$ when training model. The additional loss is as follows:
\begin{align}
    &E^* = E + r, \\
	&\mathcal{L}_{\mathrm{VAT}} \! =\! \frac{1}{n}\!\! \sum_{i=1}^{n}\!\! D_{\mathrm{KL}}\big(p(\cdot | E ; \! \Theta) \| p\left(\cdot | E^* ; \! \Theta\right)\big), \label{eq_vat_loss}
\end{align}
the adversarial perturbation $r$ is calculated by:
\begin{align}
    &E^{\prime}=E + \xi d, \label{eq_xi} \\
    &g=\nabla_{E^{\prime}} D_{\mathrm{KL}}\left(p\big(\cdot | E ; \hat{\Theta}\big) \| p\big(\cdot | E^{\prime} ; \hat{\Theta}\big)\right), \\
    &r=\epsilon g /\left\|g\right\|_{2}, \label{eq_epsilon}
\end{align}
where $\epsilon$ and $\xi$ are hyperparameters, $d$ is sampled from normal distribution $\mathcal{N}(0, I)$, $\hat{\Theta}$ is a constant set to the current parameters $\Theta$, $D_{\mathrm{KL}}(\cdot \| \cdot)$ is the KL divergence, $p(\cdot|\cdot)$ is the model conditional probability.

On the whole, the total loss of the proposed GRACE is:
\begin{align}
    \mathcal{J}(\Theta) = \mathcal{L}_e + \mathcal{L}_c + \mathcal{L}_{\mathrm{VAT}},
\end{align}
where $\mathcal{L}_e$ and $\mathcal{L}_c$ are calculated by Eq. (\ref{eq_each_weighted_loss}), denote the loss for aspect term and polarity, respectively. $\mathcal{L}_{\mathrm{VAT}}$ denotes the VAT loss, calculated by Eq. (\ref{eq_vat_loss}).

\noindent
\textbf{Consistent Polarity Label}\quad 
A question when regarding sentiment classification as polarity sequence labeling is that the generated sequence labels are not always consistent. For instance, the polarity labels may be `POS NEG' for the aspect term `operating system'. To solve this problem, we design a strategy on the representation of tokens within the same aspect term. To the generated sequence labels of ASC, we first get the boundaries of aspect terms according to the meaning of the label, e.g., the boundary of the labels `O B I O B' in Figure \ref{fig_main_structure} is $\{[1, 2), [2,4), [2,4), [4,5), [5,6)\}$, in which the element $[b_{ind}, e_{ind})$ means begin index (inclusive) and end index (exclusive). Then the aspect sentiment representation $\mathbf{G}_{c}$, and the classification is calculated as follows:
\begin{align}
    &g_i = \max (\mathbf{G}_{c}[b_{ind}:e_{ind}]), \\
    &h_i = f(g_i\mathbf{w}_h),
\end{align}
where $\mathbf{G}_{c}[b_{ind}:e_{ind}]$ is a snippet of $\mathbf{G}_{c}$ from $b_{ind}$ to $e_{ind}$ (exclusive), $\max$ is a max-pooling operator along with the sequence dimension. $\mathbf{w}_h$ is a trainable weight matrix. $f(\cdot)$ is the ReLU function. We use $h_i$ to calculate loss as Eq. (\ref{eq_softmax_pro}) and Eq. (\ref{eq_each_weighted_loss}). It is a consistent strategy to generate sentiment labels, although it cannot improve the performance in our preliminary experiments.

\section{Experiments}
\label{sec_experiments}
\subsection{Datasets}
\begin{table}[tp]
\small
\begin{center}
    \begin{tabular}{|l|l|l|l|l|}
    \hline
    Datasets & \texttt{\#POS} & \texttt{\#NEU} & \texttt{\#NEG} & \texttt{\#CON} \\ \hline \hline
    $\mathbb{D}_\text{L}$       & 1,313 & 619   & 963   & 58    \\ \hline
    $\mathbb{D}_\text{R}$       & 4,878 & 937 & 1,751 & 102 \\
    \multicolumn{1}{|r|}{- $\mathbb{D}_\text{R-14}$}   & 2,868 & 820 & 977 & 102 \\
    \multicolumn{1}{|r|}{- $\mathbb{D}_\text{R-15}$}   & 1,222 & 62  & 451 & 0   \\
    \multicolumn{1}{|r|}{- $\mathbb{D}_\text{R-16}$}   & 1,676 & 93  & 581 & 0   \\ \hline
    $\mathbb{D}_\text{T}$       & 698      & 2,254 & 271 & 0 \\ \hline
    \end{tabular}
\end{center}
\caption{\label{table_datasets} Dataset statistics. $\mathbb{D}_\text{L}$, $\mathbb{D}_\text{R}$, and $\mathbb{D}_\text{T}$ denote laptop, restaurant, and twitter datasets, respectively. \texttt{\#POS}, \texttt{\#NEU}, \texttt{\#NEG}, and \texttt{\#CON} refer to the number of positive, neutral, negative, and conflict polarity categories, respectively.}
\end{table}
We evaluate the proposed model on three benchmark sentiment analysis datasets, two of which come from the SemEval challenges, and the last comes from an English Twitter dataset, as shown in Table \ref{table_datasets}. $\mathbb{D}_\text{L}$ contains laptop reviews from SemEval 2014 \cite{Pontiki2014}, and $\mathbb{D}_\text{R}$ are restaurant reviews merged from SemEval 2014 ($\mathbb{D}_\text{R-14}$), SemEval 2015 ($\mathbb{D}_\text{R-15}$) \cite{Pontiki2015}, and SemEval 2016 ($\mathbb{D}_\text{R-16}$) \cite{Pontiki2016}. We keep the official data division of these datasets for the training set, validation set, and testing set. The reported results of $\mathbb{D}_\text{L}$ and $\mathbb{D}_\text{R}$ are average scores of 5 runs. $\mathbb{D}_\text{T}$ consists of English tweets. Due to a lack of standard train-test split, we report the ten-fold cross-validation results of $\mathbb{D}_\text{T}$ as done in \cite{Li2019a,Luo2019}. The evaluation metrics are precision (P), recall (R), and F1 score based on the exact match of aspect term and its polarity.

\subsection{Post-training}
Domain knowledge is proved useful for domain-specific tasks \cite{Xu2019,Luo2019}. In this paper, we adopt Amazon reviews~\footnote{\url{http://jmcauley.ucsd.edu/data/amazon/}} and Yelp reviews~\footnote{\url{https://www.yelp.com/academic_dataset}}, which are in-domain corpora for laptop and restaurant, respectively, to do a post-training on uncased BERT-Base for our tasks. The Amazon review dataset contains 142.8M reviews, and the Yelp review dataset contains 2.2M restaurant reviews. We combine all these reviews to finish our post-training. The maximum length of post-training is set to 320. The batch size is 4,096 for BERT-Base with gradient accumulation (every 32 steps). The BERT-Base is implemented base on the transformers library with Pytorch~\footnote{\url{https://huggingface.co/transformers}}. The mask strategy is Whole Word Masking (WWM), the same as the official BERT~\footnote{\url{https://github.com/google-research/bert}}. We use Adam optimizer and set the learning rate to be 5e-5 with 10\% warmup steps. Our pretrained model is carried out 10 epochs on 8 NVIDIA Tesla V100 GPU. We use fp16 to speed up training and to reduce memory usage. The pre-training process takes more than 5 days.

\subsection{Settings}
During fine-tuning on ATE and ASC tasks, the optimizer is Adam with 10\% warmup steps. A two-stage training strategy is utilized in our cascaded labeling model. In the first stage, we first fine-tune the ATE part initialized with the post-trained BERT weights. The learning rate is set to 3e-5 with 32 batch size, and running 5 epochs without virtual adversarial training. Then we plus virtual adversarial to continue to fine-tune 1 epoch for $\mathbb{D}_\text{L}$ and 3 epochs for other datasets with learning rate 1e-5. In the second stage, we fine-tune both ATE and ASC modules initialized with the weights from the first stage. The ASC decoder is initialized with the last corresponding layers of the ATE module. The learning rate is set to 3e-5 for the ASC part and 3e-6 for the ATE part with 32 batch size, and running 10 epochs. The maximum length is set to 128 on all datasets. $\epsilon$ in Eq. (\ref{eq_gradient_density}) is 24, and the momentum parameter $\alpha$ in Eq. (\ref{eq_momentum}) is 0.75. $\xi$ in Eq. (\ref{eq_xi}) is set to 1e-6, and $\epsilon$ in Eq. (\ref{eq_epsilon}) is set to 2. We set the shared layers $l=9$, and the number of transformer layers for ASC to 2. All the above hyper-parameters are tuned on the validation set of $\mathbb{D}_\text{L}$ and $\mathbb{D}_\text{R}$. We implement our GRACE using the same library as post-training, and all computations are done on NVIDIA Tesla V100 GPU.
\begin{table*}[tp]
    \centering
    \begin{tabular}{|p{3.0cm}|p{0.9cm}<{\centering}p{0.9cm}<{\centering}p{0.95cm}<{\centering}|p{0.9cm}<{\centering}p{0.9cm}<{\centering}p{0.95cm}<{\centering}|p{0.9cm}<{\centering}p{0.9cm}<{\centering}p{0.95cm}<{\centering}|}
    \hline
    \multicolumn{1}{|c|}{\multirow{2}{*}{Model}} & \multicolumn{3}{c|}{$\mathbb{D}_\text{L}$}                              & \multicolumn{3}{c|}{$\mathbb{D}_\text{R}$}                              & \multicolumn{3}{c|}{$\mathbb{D}_\text{T}$}                              \\ \cline{2-10} 
    \multicolumn{1}{|c|}{}                       & \multicolumn{1}{c}{P} & \multicolumn{1}{c}{R} & \multicolumn{1}{c|}{F1} & \multicolumn{1}{c}{P} & \multicolumn{1}{c}{R} & \multicolumn{1}{c|}{F1} & \multicolumn{1}{c}{P} & \multicolumn{1}{c}{R} & \multicolumn{1}{c|}{F1} \\ \hline \hline
    E2E-TBSA            & 61.27 & 54.89 & 57.90 & 68.64 & 71.01 & 69.80 & 53.08 & 43.56 & 48.01 \\
    DOER & 61.43 & 59.31 & 60.35 & 80.32 & 66.54 & 72.78 & 55.54 & 47.79 & 51.37 \\ 
    SPAN$_{Base}$       & 66.19 & 58.68 & 62.21 & 71.22 & 71.91 & 71.57 & 60.92 & 52.24 & 56.21 \\ 
    SPAN$_{Large}$      & 69.46 & 66.72 & 68.06 & 76.14 & 73.74 & 74.92 & 60.72 & 55.02 & 57.69 \\
    BERT-E2E-ABSA       & 61.88 & 60.47 & 61.12 & 72.92 & 76.72 & 74.72 & 57.63 & 54.47 & 55.94 \\ 
    \hline
    \textbf{GRACE} & 72.38 & 69.12 & \textbf{70.71} & 75.95 & 80.31 & \textbf{78.07} & 58.36 & 58.22 & \textbf{58.28} \\ 
    \multicolumn{1}{|r|}{-w/o GHL} & 68.64 & 65.90 & 67.24 & 75.16 & 78.66 & 76.87 & 55.53 & 55.62 & 55.56 \\ 
    \multicolumn{1}{|r|}{-w/o VAT} & 72.28 & 67.67 & 69.89 & 75.75 & 79.97 & 77.80 & 56.81 & 58.41 & 57.58  \\ 
    \multicolumn{1}{|r|}{-w/o PTR} & 66.39 & 61.70 & 63.96 & 73.28 & 76.53 & 74.87 & 57.26 & 58.86 & 58.04 \\ 
    \hline
    \end{tabular}
    \caption{Comparison results (\%) for aspect term-polarity pair extraction on three benchmark datasets. State-of-the-art results are marked in \textbf{bold}. `-w/o GHL' means GRACE without gradient harmonized loss, `-w/o VAT' is GRACE without virtual adversarial training, and `-w/o PTR' is GRACE without post-training on BERT-Base.}
    \label{tab_main_results}
\end{table*}
\begin{table}[htp]
    \centering
    \begin{tabular}{|l|p{0.8cm}<{\centering}|p{0.8cm}<{\centering}|p{0.8cm}<{\centering}|p{0.8cm}<{\centering}|}
        \hline
        Model     & $\mathbb{D}_\text{L}$ & $\mathbb{D}_\text{R-14}$ & $\mathbb{D}_\text{R-15}$ & $\mathbb{D}_\text{R-16}$ \\ \hline \hline
        IMN & 58.37 & 69.54 & 59.18 & - \\
        DREGCN & 63.04 & 72.60 & 62.37 & - \\
        WHW & 62.34 & 71.95 & 65.79 & 71.73 \\
        TAS-BERT & - & - & 66.11 & 75.68 \\
        IKTN-BERT & 62.34 & 71.75 & 62.33 & - \\
        DHGNN & 59.61 & 68.91 & 58.37 & - \\ 
        RACL-BERT & 63.40  & 75.42 & 66.05 & - \\ \hline
        \textbf{GRACE} & \textbf{70.71} & \textbf{77.26} & \textbf{68.16} & \textbf{76.49} \\
        \multicolumn{1}{|r|}{-w/o GHL} & 67.24  & 75.83 & 66.73 & 75.09  \\
        \multicolumn{1}{|r|}{-w/o VAT} & 69.89 & 77.16 & 67.75 & 76.03   \\
        \multicolumn{1}{|r|}{-w/o PTR} & 63.96 & 71.56 & 59.82 & 66.95  \\
        \hline
    \end{tabular}
    \caption{Comparison results of F1 score (\%) for aspect term-polarity pair extraction on four benchmark datasets. `-' denotes unreported results. `-w/o GHL', `-w/o VAT', and `-w/o PTR' have the same meaning as which in Table \ref{tab_main_results}.}
    \label{tab_main_results_noimp}
\end{table}

\subsection{Baseline Methods}
We compare our model~\footnote{Code and pre-trained weights will be released at: \url{https://github.com/ArrowLuo/GRACE}} with the following models:

\noindent
\textbf{E2E-TBSA} \cite{Li2019a} is an end-to-end model of the collapsed approach proposed to address ATE and ASC simultaneously.

\noindent
\textbf{DOER} \cite{Luo2019} employs a cross-shared unit to train the ATE and ASC jointly.

\noindent
\textbf{SPAN} \cite{Hu2019} is a pipeline approach built on BERT-Large (SPAN$_{Large}$) to solve aspect term-sentiment pairs extraction. We implement its BERT-Base version (SPAN$_{Base}$) using the available code~\footnote{\url{https://github.com/huminghao16/SpanABSA}}.

\noindent
\textbf{BERT-E2E-ABSA} \cite{Li2019b} is a BERT-based benchmark for aspect term-sentiment pairs extraction. We use the BERT+GRU for $\mathbb{D}_\text{L}$ and BERT+SAN for $\mathbb{D}_\text{R}$ as our baselines due to their best-reported performance. Besides, we produce the results on $\mathbb{D}_\text{T}$ with BERT+SAN keeping the settings the same as on $\mathbb{D}_\text{R}$~\footnote{\url{https://github.com/lixin4ever/BERT-E2E-ABSA}}.

We compare our model with the above baselines on $\mathbb{D}_\text{L}$, $\mathbb{D}_\text{R}$, and $\mathbb{D}_\text{T}$, and compare it with the following baselines on $\mathbb{D}_\text{L}$, $\mathbb{D}_\text{R-14}$, $\mathbb{D}_\text{R-15}$, and $\mathbb{D}_\text{R-16}$ because of the common datasets reported by the official implementation.

\noindent
\textbf{IMN} \cite{He2019} uses an interactive architecture with multi-task learning for end-to-end ABSA tasks. It contains aspect term and opinion term extraction besides aspect-level sentiment classification.

\noindent
\textbf{DREGCN} \cite{Liang2020a} designs a dependency syntactic knowledge augmented interactive architecture with multi-task learning for end-to-end ABSA. DREGCN is short for the official DREGCN+CNN+BERT due to its better performance.

\noindent
\textbf{WHW} \cite{Peng2020} develops a two-stage framework to address aspect term extraction, aspect sentiment classification,
and opinion extraction.

\noindent
\textbf{TAS-BERT} \cite{Wan2020} proposes a method based on BERT-Base that can capture the dependence on both aspect terms and categories for sentiment prediction. TAS-BERT is short for the official TAS-BERT-SW-BIO-CRF due to its better performance.

\noindent
\textbf{IKTN+BERT} \cite{Liang2020} discriminately transfers the document-level linguistic knowledge to aspect term, opinion term extraction, and aspect-level sentiment classification.

\noindent
\textbf{DHGNN} \cite{Liu2020} presents a dynamic heterogeneous graph to model the aspect extraction and sentiment detection explicitly jointly.

\noindent
\textbf{RACL-BERT} \cite{chen2020racl} is built on BERT-Large and allows the aspect term extraction, opinion term extraction, and aspect-level sentiment classification to work coordinately via the multi-task learning and relation propagation mechanisms in a stacked multi-layer network.

\subsection{Results and Analysis}
\noindent
\textbf{Comparison Results.}\quad The comparison results are shown in Table \ref{tab_main_results} and Table \ref{tab_main_results_noimp} because different baselines officially report on different datasets. Overall, our proposed GRACE consistently obtains the best F1 score across all datasets and significantly outperforms the strongest baselines in most cases on aspect term-polarity co-extraction. Compared to the state-of-the-art pipeline approach, the GRACE outperforms SPAN$_{Base}$ by 8.50\%, 6.50\%, and 2.07\% on $\mathbb{D}_\text{L}$, $\mathbb{D}_\text{R}$, and $\mathbb{D}_\text{T}$, respectively. Even comparing to SPAN$_{Large}$ built on 24-layers BERT-Large, the improvements are still 2.65\%, 3.15\%, and 0.59\% on $\mathbb{D}_\text{L}$, $\mathbb{D}_\text{R}$, and $\mathbb{D}_\text{T}$, respectively. It indicates that a carefully-designed joint model has capable of achieving better performance than pipeline approaches on our task. Compared to other multi-task models containing additional information, e.g., opinion terms and aspect term categories, the GRACE achieves absolute gains over the IMN, WHW, TAS-BERT, IKTN+BERT, and RACL-BERT at least by 7.31\%, 1.84\%, 2.05\%, and 0.81\% on $\mathbb{D}_\text{L}$, $\mathbb{D}_\text{R-14}$, $\mathbb{D}_\text{R-15}$, $\mathbb{D}_\text{R-16}$, respectively. It suggests that GRACE can extend to more tasks of ABSA.

\vspace{+1mm}
\noindent
\textbf{Ablation Study.}\quad To study the effectiveness of the gradient harmonized loss (GHL), VAT, and post-pretraining, we conduct ablation experiments on each of them. The results are shown in the second block in Table \ref{tab_main_results} and Table \ref{tab_main_results_noimp}. We can see that the scores drop more seriously without GHL comparing to that without VAT. It points out that GRACE can benefit more from the gradient harmonized loss than VAT, and alleviate the imbalance issue of labels is more important to the sequence labeling. The drop of scores without post-training is the worst on all laptop and restaurant datasets, which indicates that the domain-specific knowledge can improve the task-related datasets massively.
\begin{table}[tp]
    \begin{center}
        \begin{tabular}{|l|p{1.1cm}<{\centering}|p{1.1cm}<{\centering}|p{1.1cm}<{\centering}|}
            \hline
            Model     & $\mathbb{D}_\text{L}$ & $\mathbb{D}_\text{R}$ & $\mathbb{D}_\text{T}$ \\ \hline \hline
            DE-CNN             & 81.26 & 78.98 & 63.23  \\ 
            DOER               & 82.61 & 81.06 & 71.35 \\ 
            SPAN$_{Large}$     & 83.35 & 82.38 & 75.28 \\ 
            BERT-PT            & 84.26 & - & - \\ 
            BERT-PT-AUG        & 85.33 & - & - \\ 
            BAT                & 85.57 & - & - \\ 
            \hline
            \textbf{GRACE}      & \textbf{87.93} & \textbf{85.45} & \textbf{75.73} \\ 
            \multicolumn{1}{|r|}{-w/o ASC} & 87.45 & 84.49 & 75.52 \\ 
            \hline
            \end{tabular}
    \end{center}
    \caption{\label{table_results_ate} F1 score (\%) comparison for aspect term extraction. `-' denotes unreported results. `-w/o ASC' means training without the ASC branch.}
\end{table}
\begin{table*}[tp]
    \small
	\begin{center}
		\begin{tabular}{m{5.8cm}|p{2.7cm}<{\centering}|p{2.7cm}<{\centering}|p{2.7cm}<{\centering}}
			\hline
            \textbf{Sentence} & \multicolumn{1}{c|}{\textbf{BASE}} & \multicolumn{1}{c|}{\textbf{GRACE w/o GHL}} & \multicolumn{1}{c}{\textbf{GRACE}} \\ \hline \hline
            
            \multirow{3}{5.8cm}{I used [\textbf{windows XP}]$_\texttt{NEU}$, [\textbf{windows Vista}]$_\texttt{NEU}$, and [\textbf{Windows 7}]$_\texttt{NEU}$ extensively.} 
            & [windows XP]$_\texttt{POS}$ (\xmark) & [windows XP]$_\texttt{NEU}$ & [windows XP]$_\texttt{NEU}$ \\
            & [windows Vista]$_\texttt{NEU}$ & [windows Vista]$_\texttt{NEU}$ & [windows Vista]$_\texttt{NEU}$ \\
            & [Windows 7]$_\texttt{NEU}$ & [Windows 7]$_\texttt{NEU}$ & [Windows 7]$_\texttt{NEU}$ \\ \hline
            
            \multirow{2}{5.8cm}{User upgradeable [\textbf{RAM}]$_\texttt{POS}$ and [\textbf{HDD}]$_\texttt{POS}$.}
             & [RAM]$_\texttt{POS}$ & [RAM]$_\texttt{POS}$ & [RAM]$_\texttt{POS}$ \\ 
             & [HDD]$_\texttt{NEU}$ (\xmark) & [HDD]$_\texttt{POS}$ & [HDD]$_\texttt{POS}$ \\ \hline 
            
            Although somewhat loud, the [\textbf{noise}]$_\texttt{CON}$ was minimally intrusive. & [noise]$_\texttt{POS}$ (\xmark) & [noise]$_\texttt{POS}$ (\xmark) & [noise]$_\texttt{CON}$  \\ \hline

            The [\textbf{atmosphere}]$_\texttt{CON}$ was nice but it was a little too dark. & [atmosphere]$_\texttt{POS}$ (\xmark) & [atmosphere]$_\texttt{POS}$ (\xmark) & [atmosphere]$_\texttt{CON}$ \\ \hline
		\end{tabular}
	\end{center}
	\caption{\label{table_case_study} Case analysis on BASE, GRACE w/o GHL, and GRACE. \ding{55} means wrong prediction.}
\end{table*}

\vspace{+1mm}
\noindent
\textbf{Results on ATE.}\quad As an extra output of the proposed GRACE, we also compare ATE results with state-of-the-art baselines. \textbf{DE-CNN} \cite{Xu2018} adopts CNN training on general purpose embeddings  domain specific embeddings to finish ATE. \textbf{BERT-PT} \cite{Xu2019} post-trains BERT’s weights using in-domain review datasets and MRC dataset. It is implemented based on BERT-Base. \textbf{BERT-PT-AUG} \cite{Li2020} is an improvement version of BERT-PT with a controllable data augmentation approach. \textbf{BAT} \cite{Karimi2020} is a BERT adversarial training model. The results of the ATE are shown in Table \ref{table_results_ate}. Our GRACE achieves state-of-the-art results over baselines. The lower scores of GRACE without the ASC branch indicate that the ASC task could enhance the ATE.

\vspace{+1mm}
\noindent
\textbf{Results on Cascaded Labeling.}\quad To verify the effectiveness of our cascaded labeling strategy, as a particular case of the GRACE, we set the shared layers $l=12$ and set the number of transformer layers for ASC to 0, and refer it as \textbf{BASE}. Thus, there is no generated aspect term label from ATE branch when training the ASC branch. The F1 scores of BASE are 68.35\% and 76.76\% on $\mathbb{D}_\text{L}$ and $\mathbb{D}_\text{R}$, respectively. The results are lower than 70.71\% and 78.07\% of GRACE on the same datasets. This fact indicates that considering the interaction between aspect terms and paying more attention to other tokens are benefit to the sentiment labeling.

\vspace{+1mm}
\noindent
\textbf{Case Study.}\quad Table \ref{table_case_study} shows some examples of BASE, GRACE without gradient harmonized loss (w/o GHL), and GRACE sampled from $\mathbb{D}_\text{L}$ and $\mathbb{D}_\text{R}$. As observed in the first two examples, the GRACE  incorrectly predicts both aspect terms and their sentiments. Comparing with the BASE, we believe the cascaded labeling strategy can make an interaction between aspect terms within a sentence, which enhances the judgment of sentiment labels. The last two rows indicate that GRACE can get correct results, even the \texttt{CON} is minimal. The reason is not only the more comprehensive information proved by cascaded labeling strategy but also the balance of labels given by gradient harmonized loss.

\section{Related Work}
\label{sec_related_work}
Aspect term extraction and aspect sentiment classification are two major topics of aspect-based sentiment analysis. Many researchers have studied each of them for a long time. For the ATE task, unsupervised methods such as frequent pattern mining \cite{Hu2004}, rule-based approach \cite{Qiu2011, Liu2015Automated}, topic modeling \cite{He2011auto, Chen2014Aspect}, and supervised methods such as sequence labeling based models \cite{Wang2016a,Yin2016,Xu2018,Li2018a,Luo2019a,Ma2019} are two main directions. For the ASC task, the relation or position between the aspect terms and the surrounding context words are usually used \cite{Tang2016a,Arjun2016extract}. Besides, there are some other approaches, such as convolution neural networks \cite{Poria2016aspect,Li2018}, attention networks \cite{Wang2016attention,Ma2017a,He2017}, memory networks \cite{Wang2018a}, capsule network \cite{Chen2019}, and graph neural networks \cite{Wang2020}.

We regard ATE and ASC as two parallel sequence labeling tasks in this paper. Compared with the separate methods, this approach can concisely generate all aspect term-polarity pairs of input sentences. Like our work, \citet{Mitchell2013} and \citet{Zhang2015} are also about performing two sequence labeling tasks, but they extract named entities and their sentiment classes jointly. We have a different objective and utilize a different model. \citet{Li2017}, \citet{ma-etal-2018-joint} and \citet{Li2019a} have the same objective as us. The main difference is that their approaches belong to a collapsed approach, but ours is a joint approach. \citet{Luo2019} use joint approach like ours, they focus on the interaction between two tasks, and some extra objectives are designed to assist the extraction. \citet{Hu2019} consider the ATE as a span extraction question, and extract aspect term and its sentiment polarity using a pipeline approach. There are some other approaches to address these two tasks \cite{Li2019b,He2019,Liang2020a,Peng2020,Wan2020,Liang2020,Liu2020,chen2020racl}. However, almost all of previous models do not concern the imbalance of labels in such sequence labeling tasks. To the best of our knowledge, this is the first work to alleviate the imbalance issue in the ABSA.

\section{Conclusion}
\label{sec_conclusion}
In this paper, we proposed a novel framework GRACE to solve aspect term extraction and aspect sentiment classification simultaneously. The proposed framework adopted a cascaded labeling approach to enhance the interaction between aspect terms and improve the attention of sentiment tokens for each term by a multi-head attention architecture. Besides, we alleviated the imbalance issue of labels in our labeling tasks by a gradient harmonized method borrowed from object detection. The virtual adversarial training and post-training on domain datasets were also introduced to improve the extraction performance. Experimental results on three benchmark datasets verified the effectiveness of GRACE and showed that it significantly outperforms the baselines on aspect term-polarity co-extraction.

\section*{Acknowledgments}
This work was supported by National Key R\&D Program of China (2019YFB2101802) and Sichuan Key R\&D project (2020YFG0035).

\bibliographystyle{acl_natbib}
\bibliography{ImbalanceSequenceLabellingRef}

\end{document}